\documentclass{article}
\usepackage{ijcai20}

\usepackage{times}
\usepackage{soul}
\usepackage{url}
\usepackage[hidelinks]{hyperref}
\usepackage[utf8]{inputenc}
\usepackage[small]{caption}
\usepackage{graphicx}
\usepackage{amsmath}
\usepackage{amsthm}
\usepackage{booktabs}
\usepackage{algorithm}
\usepackage{algorithmic}
\usepackage{amsfonts}
\usepackage{amsmath}
\usepackage{multirow, makecell}
\usepackage{graphicx}
\usepackage{enumitem}
\usepackage{textcomp}
\usepackage{xcolor}
\usepackage{subcaption}
\usepackage{esvect}

\newcommand{\action}[2]{$\rm\scriptstyle {#1}\to\rm{#2}$}
\newcommand*{\affaddr}[1]{#1}
\newcommand*{\affmark}[1][*]{$^#1$}
\newcommand*{\email}[1]{#1}

\usepackage{latexsym} 

\newcommand{\citet}[1] {\citeauthor{#1}~\shortcite{#1}}
\def\doubleunderline#1{\underline{\underline{#1}}}

\title{How Far are We from Effective Context Modeling?\\An Exploratory Study on Semantic Parsing in Context}

\author{
    Qian Liu\affmark[1]{\thanks{Work done during an internship at Microsoft Research.}},
    Bei Chen\affmark[2],
    Jiaqi Guo\affmark[3]$^*$,
    Jian-Guang Lou\affmark[2],
    \textbf{Bin Zhou\affmark[1]\And
    Dongmei Zhang\affmark[2]}
    \affiliations
    \affaddr{\affmark[1]School of Computer Science and Engineering, Beihang University, China}\\
    \affaddr{\affmark[2]Microsoft Research, Beijing, China}\\
    \affaddr{\affmark[3]Xi'an Jiaotong University, Xi'an, China}\\
    \emails
    \email{\{qian.liu, zhoubin\}@buaa.edu.cn, \{beichen, jlou, dongmeiz\}@microsoft.com},\\
    \email{jasperguo2013@stu.xjtu.edu.cn}
}

\begin{document}

\maketitle

\begin{abstract}
Recently semantic parsing in context has received considerable attention, which is challenging since there are complex contextual phenomena. Previous works verified their proposed methods in limited scenarios, which motivates us to conduct an exploratory study on context modeling methods under real-world semantic parsing in context. We present a grammar-based decoding semantic parser and adapt typical context modeling methods on top of it. We evaluate $13$ context modeling methods on two large complex cross-domain datasets, and our best model achieves state-of-the-art performances on both datasets with significant improvements. Furthermore, we summarize the most frequent contextual phenomena, with a fine-grained analysis on representative models, which may shed light on potential research directions. Our code is available at https://github.com/microsoft/ContextualSP.

\end{abstract}

\section{Introduction}\label{sec:intro}

\begin{figure}[t]
    \centering
    \includegraphics[width=0.92\linewidth]{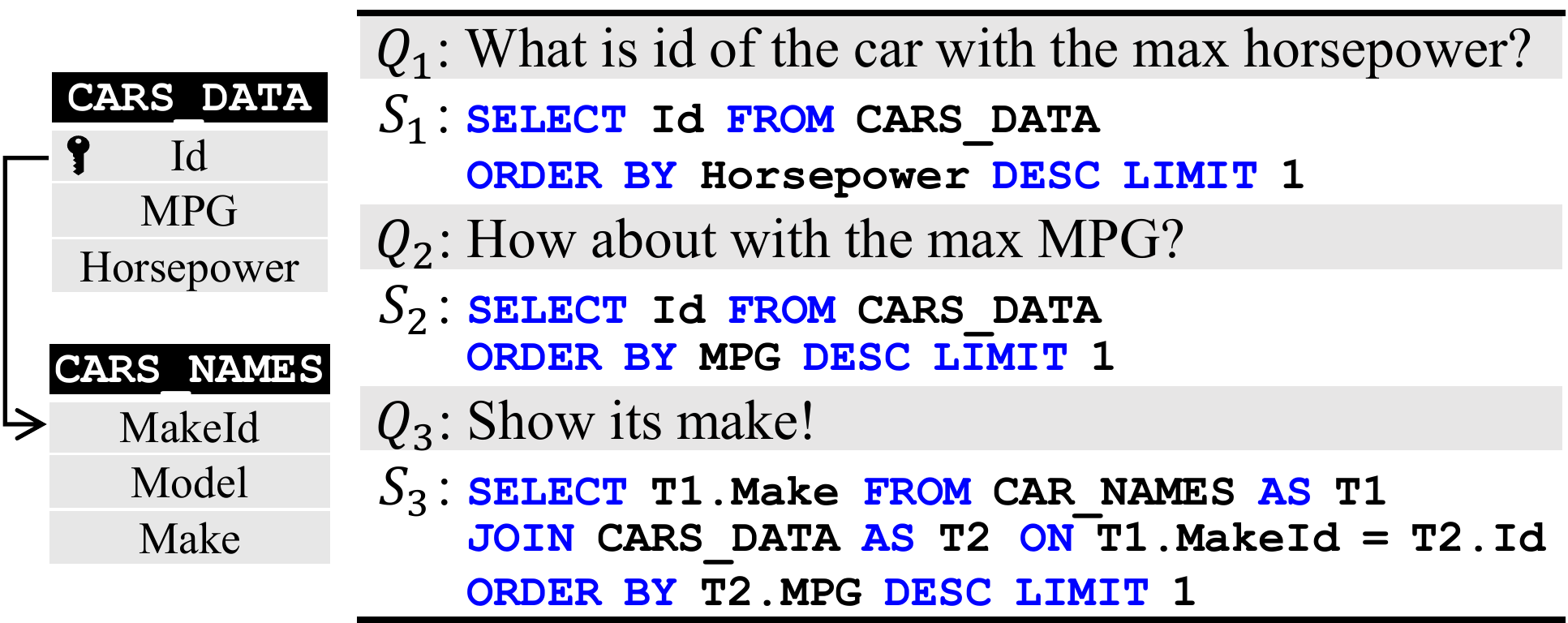}
    \caption{An example dialogue (right) and its database schema (left).}
    \label{fig:context_intro}
\end{figure}

Semantic parsing, which translates a natural language sentence into its corresponding executable logic form (e.g. Structured Query Language, \textbf{SQL}), relieves users from the burden of learning techniques behind the logic form. The majority of previous studies on semantic parsing assume that queries are context-independent and analyze them in isolation. However, in reality, users prefer to interact with systems in a dialogue, where users are allowed to ask context-dependent incomplete questions \cite{bertomeu-etal-2006-contextual}. That arises the task of Semantic Parsing in Context (\textbf{SPC}), which is quite challenging as there are complex contextual phenomena. In general, there are two sorts of contextual phenomena in dialogues: \textit{Coreference} and \textit{Ellipsis} \cite{androutsopoulos1995natural}. Figure~\ref{fig:context_intro} shows a dialogue from the dataset \textsc{SParC}~\cite{yu-etal-2019-sparc}. After the question ``What is id of the car with the max horsepower?'', the user poses an elliptical question ``How about with the max MPG?'', and a question containing pronouns ``Show its make!''. Only when completely understanding the context, could a parser successfully parse the incomplete questions into their corresponding SQL queries.

A number of context modeling methods have been suggested in the literature to address SPC~\cite{iyyer-etal-2017-search,suhr-etal-2018-learning,yu-etal-2019-sparc,zhang-etal-2019-editing,yu-etal-2019-cosql}. These methods proposed to leverage two categories of context: recent questions and precedent logic form. It is natural to leverage recent questions as context. Taking the example from Figure~\ref{fig:context_intro}, when parsing $Q_3$, we also need to take $Q_1$ and $Q_2$ as input. We can either simply concatenate the input questions, or use a model to encode them hierarchically \cite{suhr-etal-2018-learning}. As for the second category, instead of taking a sequence of recent questions as input, it only considers the precedent logic form. For instance, when parsing $Q_3$, we only need to take $S_2$ as context. With such a context, the decoder can attend over it, or reuse it via a copy mechanism \cite{suhr-etal-2018-learning,zhang-etal-2019-editing}. Intuitively, methods that fall into this category enjoy better generalizability, as they only rely on the last logic form as context, no matter at which turn. Notably, these two categories of context can be used simultaneously.

However, it remains unclear how far we are from effective context modeling. First, there is a lack of thorough comparisons of typical context modeling methods on complex SPC (e.g. cross-domain). Second, none of previous works verified their proposed context modeling methods with the grammar-based decoding technique, which has been proven to be highly effective in semantic parsing \cite{krishnamurthy-etal-2017-neural,yin18emnlpdemo,guo-etal-2019-towards,lin2019grammarbased}. To obtain better performance, it is worthwhile to study how context modeling methods collaborate with the grammar-based decoding. Last but not least, there is a limited understanding of how context modeling methods perform on various contextual phenomena. An in-depth analysis can shed light on potential research directions.

In this paper, we try to fulfill the above insufficiency via an exploratory study on real-world semantic parsing in context. Concretely, we present a grammar-based decoding semantic parser and adapt typical context modeling methods on top of it. Through experiments on two large complex cross-domain datasets, \textsc{SParC} \cite{yu-etal-2019-sparc} and \textsc{CoSQL} \cite{yu-etal-2019-cosql}, we carefully compare and analyze the performance of different context modeling methods. Our best model achieves state-of-the-art (SOTA) performances on both datasets with significant improvements. Furthermore, we summarize and generalize the most frequent contextual phenomena, with a fine-grained analysis of representative models. Through the analysis, we obtain some interesting findings, which may benefit the community on the potential research directions.

\section{Methodology}\label{sec:method}

In the task of SPC, we are given a dataset composed of dialogues. Denoting $\langle \mathbf{x}_1,...,\mathbf{x}_n\rangle$ a sequence of natural language questions in a dialogue, $\langle\mathbf{y}_1,...,\mathbf{y}_n\rangle$ are their corresponding SQL queries. Each SQL query is conditioned on a multi-table database schema, and the databases used in test do not appear in training. In this section, we first present a base model without considering context. Then we introduce $6$ typical context modeling methods and describe how we equip the base model with these methods. Finally, we present how to augment the model with BERT~\cite{devlin-etal-2019-bert}.

\subsection{Base Model}\label{sec:base_model}

We employ the popularly used attention-based sequence-to-sequence architecture \cite{sutskever2014sequence,Bahdanau2014NeuralMT} to build our base model. As shown in Figure~\ref{fig:model_overview}, the base model consists of a question encoder and a grammar-based decoder. For each question, the encoder provides contextual representations, while the decoder generates its corresponding SQL query according to a predefined grammar. 

\subsubsection{Question Encoder}

To capture contextual information within a question, we apply Bidirectional Long Short-Term Memory Neural Network (BiLSTM) as our question encoder \cite{hochreiter1997long,schuster1997bidirectional}. Specifically, at turn $i$, firstly every token $x_{i,k}$ in $\mathbf{x}_{i}$ is fed into a word embedding layer $\mathbf{\phi}^x$ to get its embedding representation $\mathbf{\phi}^x{(x_{i,k})}$. On top of the embedding representation, the question encoder obtains a contextual representation $\mathbf{h}^{E}_{i,k}=[\mathop{{\mathbf{h}}^{\overrightarrow{E}}_{i,k}}\,;{\mathbf{h}}^{\overleftarrow{E}}_{i,k}]$, where the forward hidden state is computed as following:
\begin{equation}\label{eq:for_ques_encode}
\mathbf{h}^{\overrightarrow{E}}_{i,k} = {\mathbf{LSTM}^{\overrightarrow{E}}} \Big( \mathbf{\phi}^x{(x_{i,k})},\mathbf{h}^{\overrightarrow{E}}_{i,k-1} \Big).
\end{equation}

\subsubsection{Grammar-based Decoder}

\begin{figure}[t]
    \centering
    \includegraphics[width=.45\textwidth]{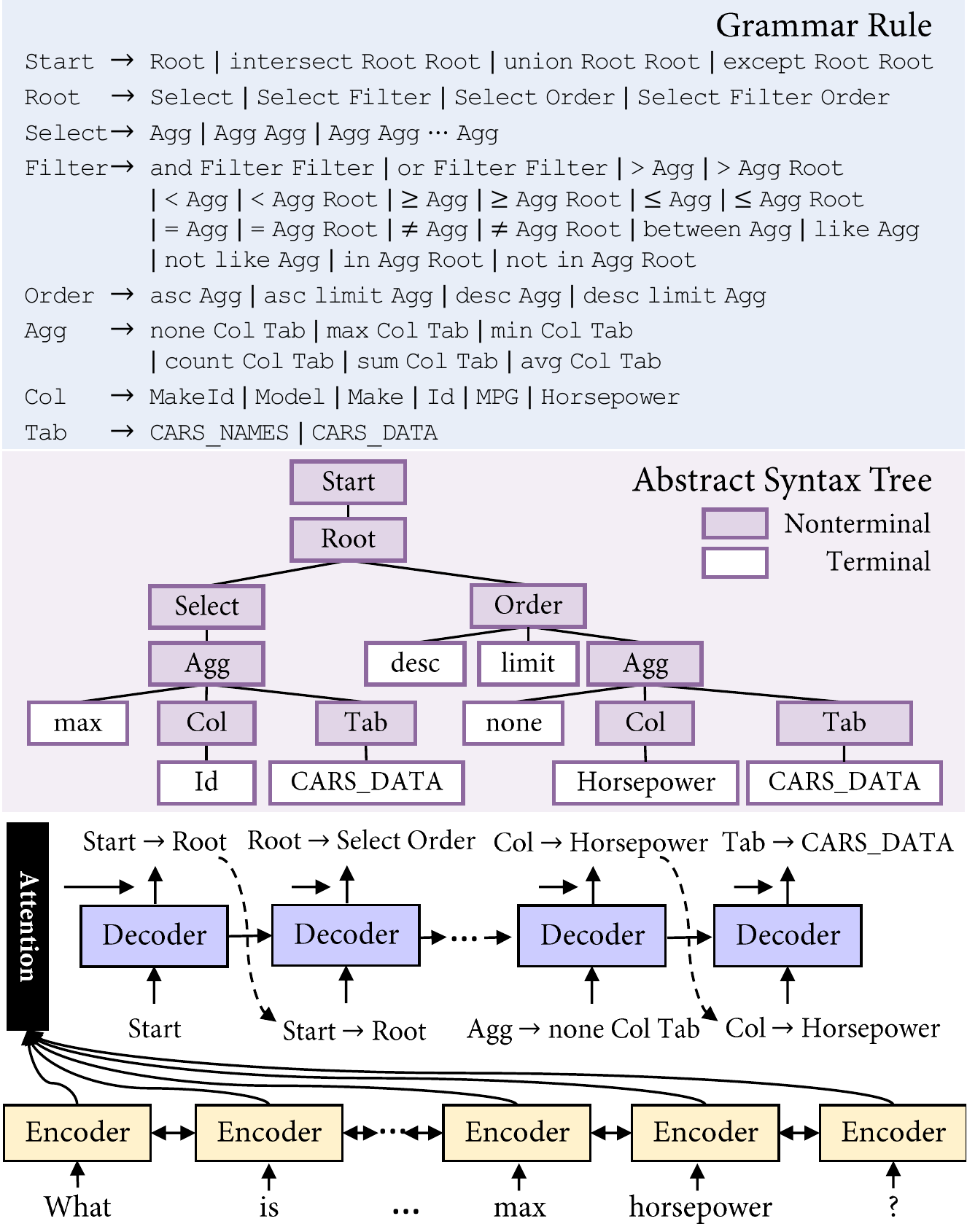}
    \caption{The grammar rule and the abstract syntax tree for the SQL \textbf{\texttt{{\color{blue}SELECT} Id {\color{blue}FROM} CARS\_DATA {\color{blue}ORDER BY} Horsepower {\color{blue}DESC LIMIT} 1}}, along with the framework of our base model. Schema-specific grammar rules change with the database schema.}
    \label{fig:model_overview}
\end{figure}

The decoder is grammar-based with attention on the input question \cite{krishnamurthy-etal-2017-neural}. Different from producing a SQL query word by word, our decoder outputs a sequence of \textit{grammar rule} (i.e. action). Such a sequence has one-to-one correspondence with the abstract syntax tree of the SQL query. Taking the SQL query in Figure~\ref{fig:model_overview} as an example, it is transformed to the action sequence $\langle$ \action{Start}{Root}, \action{Root}{Select\ Order}, \action{Select}{Agg}, \action{Agg}{max\ Col\ Tab}, \action{Col}{Id}, \action{Tab}{CARS\_DATA}, \action{Order}{desc\ limit\ Agg}, \action{Agg}{none\ Col\ Tab}, \action{Col}{Horsepower}, \action{Tab}{CARS\_DATA} $\rangle$ by left-to-right depth-first traversing on the tree. At each decoding step, a \textit{nonterminal} is expanded using one of its corresponding grammar rules. The rules are either schema-specific (e.g.\,\action{Col}{Horsepower}), or schema-agnostic (e.g.\,\action{Start}{Root}). More specifically, as shown at the top of Figure \ref{fig:model_overview}, we make a little modification on $\rm{Order}$-related rules upon the grammar proposed by \citet{guo-etal-2019-towards}, which has been proven to have better performance than vanilla SQL grammar. Denoting $\mathbf{LSTM}^{\overrightarrow{D}}$ the unidirectional LSTM used in the decoder, at each decoding step $j$ of turn $i$, it takes the embedding of the previous generated grammar rule $\mathbf{\phi}^y(y_{i,j-1})$ (indicated as the dash lines in Figure~\ref{fig:model_overview}), and updates its hidden state as:
\begin{equation}\label{eq:decoder}
\mathbf{h}^{\overrightarrow{D}}_{i,j} = {\mathbf{LSTM}^{\overrightarrow{D}}} \Big( [\mathbf{\phi}^y(y_{i,j-1});\mathbf{c}_{i,j-1}],\mathbf{h}^{\overrightarrow{D}}_{i,j-1} \Big),
\end{equation}
where $\mathbf{c}_{i,j-1}$ is the context vector produced by attending on each encoder hidden state $\mathbf{h}^E_{i,k}$ in the previous step:
\begin{equation}\label{eq:attn}
  \begin{aligned}
    e_{i,k}=\mathbf{h}^E_{i,k}\;\mathbf{W}^e\;\mathbf{h}^{\overrightarrow{D}}_{i,j-1}&,\ a_{i,k} = \frac{\exp{(e_{i,k})}}{\sum_{k} \exp({e_{i,k}})}, \\
    \mathbf{c}_{i,j-1}&=\sum\nolimits_{k} \mathbf{h}^E_{i,k} \cdot a_{i,k},
  \end{aligned}
\end{equation}
where $\mathbf{W}^e$ is a learned matrix. $\mathbf{h}^{\overrightarrow{D}}_{i,0}$ is initialized by the final encoder hidden state $\mathbf{h}^E_{i,|\mathbf{x}_{i}|}$, while $\mathbf{c}_{i,0}$ is a zero-vector. For each schema-agnostic grammar rule, $\mathbf{\phi}^y$ returns a learned embedding. The embedding of a schema-specific grammar rule is obtained by passing its schema (i.e. table or column) through another unidirectional LSTM, namely schema encoder $\mathbf{LSTM}^{\overrightarrow{S}}$. For example, the embedding of \action{Col}{Id} is:
\begin{equation}
    \mathbf{\phi}^y({\scriptstyle {\rm Col}\to{\rm Id}})=\mathbf{LSTM}^{\overrightarrow{S}}\big(\mathbf{\phi}^x(\text{``${\rm Id}$''}),{\vv{\mathbf{0}}}\big).
\end{equation}

As for the output $y_{i,j}$, if the expanded nonterminal corresponds to schema-agnostic grammar rules, we can obtain the output probability of action ${\gamma}$ as:
\begin{equation}\label{eq:global_output}
    P(y_{i,j}={\gamma})\ {\propto}\ \exp\big(\tanh{([\mathbf{h}^{\overrightarrow{D}}_{i,j};\mathbf{c}_{i,j}]\,\mathbf{W}^o)}\,\mathbf{\phi}^y({\gamma})\big),
\end{equation}
where $\mathbf{W}^o$ is a learned matrix. When it comes to schema-specific grammar rules, the main challenge is that the model may encounter schemas never appeared in training due to the cross-domain setting. To deal with it, we do not directly compute the similarity between the decoder hidden state and the schema-specific grammar rule embedding. Instead, we first obtain the unnormalized linking score $l(x_{i,k},\gamma)$ between the $k$-th token in $\mathbf{x}_i$ and the schema inside action $\gamma$. It is computed by both handcraft features (e.g. word exact match) \cite{bogin-etal-2019-representing} and learned similarity (i.e. dot product between word embedding and grammar rule embedding). With the input question as bridge, we reuse the attention score $a_{i,k}$ in Equation \ref{eq:attn} to measure the probability of outputting a schema-specific action $\gamma$ as:
\begin{equation}\label{eq:schema_output}
    P(y_{i,j}={\gamma})\ {\propto}\ \exp\big(\sum\nolimits_{k} a_{i,k}\!\cdot\!l(x_{i,k}, \gamma)\big).
\end{equation}

\subsection{Recent Questions as Context}\label{sec:question_history}

\begin{figure}[t]
    \centering
    \includegraphics[width=0.7\linewidth]{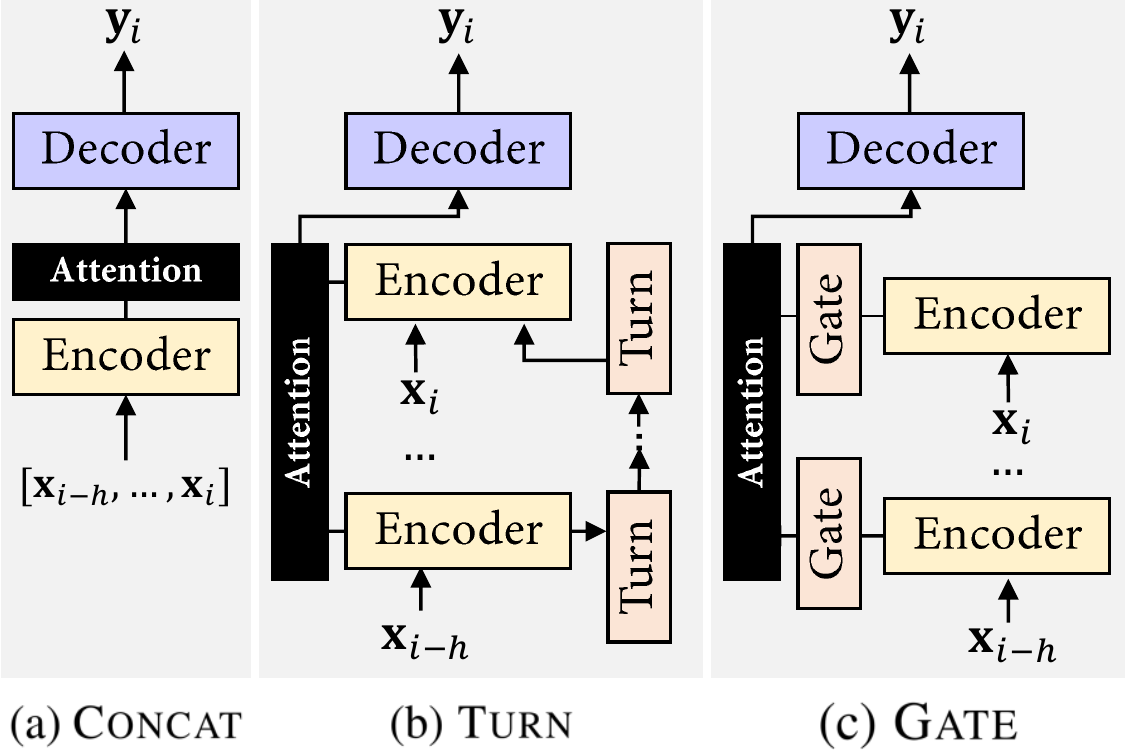}
    \caption{Different methods to incorporate recent $h$ questions $[ \mathbf{x}_{i-h},...,\mathbf{x}_{i-1}]$. (a) \textsc{Concat}: concatenate recent questions with $\mathbf{x}_{i}$ as input; (b) \textsc{Turn}: employ a turn-level encoder to capture the inter-dependencies among questions in different turns; (c) \textsc{Gate}: use a gate mechanism to compute the importance of each question.}
    \label{fig:diff_utt_context}
\end{figure}

To take advantage of the question context, we provide the base model with recent $h$ questions as additional input. As shown in Figure~\ref{fig:diff_utt_context}, we summarize and generalize three ways to incorporate recent questions as context.

\paragraph{\textsc{Concat}.} The method concatenates recent questions with the current question in order, making the input of the question encoder be $[\mathbf{x}_{i-h},\dots,\mathbf{x}_{i}]$, while the architecture of the base model remains the same. We do not insert special delimiters between questions, as there are punctuation marks.

\paragraph{\textsc{Turn}.} A dialogue can be seen as a sequence of questions which, in turn, are sequences of words. Considering such hierarchy, \citet{suhr-etal-2018-learning} employed a turn-level encoder (i.e. an unidirectional LSTM) to encode recent questions hierarchically. At turn $i$, it takes the previous question vector $\mathbf{h}^{E}_{i-1}=[\mathbf{h}^{\overleftarrow{E}}_{i-1,1},\mathbf{h}^{\overrightarrow{E}}_{i-1,|\mathbf{x}_{i-1}|}]$ as input, and updates its hidden state to $\mathbf{h}^{\overrightarrow{T}}_{i}$. Then $\mathbf{h}^{\overrightarrow{T}}_{i}$ is fed into the question encoder as an implicit context. Accordingly Equation~\ref{eq:for_ques_encode} is rewritten:
\begin{equation}
\mathbf{h}^{\overrightarrow{E}}_{i,k} = {\mathbf{LSTM}^{\overrightarrow{E}}} \Big( [\mathbf{\phi}^x{(x_{i,k})};\mathbf{h}^{\overrightarrow{T}}_{i}],\mathbf{h}^{\overrightarrow{E}}_{i,k-1} \Big).
\end{equation}
Similar to \textsc{Concat}, \citet{suhr-etal-2018-learning} allowed the decoder to attend over all encoder hidden states. To make the decoder distinguish hidden states from different turns, they further proposed a relative distance embedding ${\phi}^{d}$ in attention computing. Taking the above into account, Equation~\ref{eq:attn} is as:
\begin{equation}\label{eq:inter_attn}
    \begin{aligned}
    e_{i-t,k}&=[\mathbf{h}^E_{i-t,k};\mathbf{\phi}^d(t)]\;\mathbf{W}^e\;\mathbf{h}^{\overrightarrow{D}}_{i,j-1},\\
    a_{i-t,k}&= \frac{\exp{(e_{i-t,k})}}{\sum_{t}\sum_{k} \exp({e_{i-t,k}})}, \\
    \mathbf{c}_{i,j-1}&=\sum\nolimits_{t}\sum\nolimits_{k} [\mathbf{h}^E_{i-t,k};\mathbf{\phi}^d(t)] \cdot a_{i-t,k},
    \end{aligned}
\end{equation}
where $t{\in}[0,\dots,h]$ represents the relative distance.

\paragraph{\textsc{Gate}.} To jointly model the decoder attention in token-level and question-level, inspired by the advances of open-domain dialogue area \cite{zhang-etal-2018-context}, we propose a gate mechanism to automatically compute the importance of each question. The importance is computed by:
\begin{equation}
    \begin{aligned}
    g_{i-t} &= \mathbf{V}^{g}\tanh(\mathbf{U}^g\mathbf{h}^{E}_{{i-t}}+\mathbf{W}^g\mathbf{h}^{E}_{i}), \\
    \bar{g}_{i-t} &= \frac{\exp(g_{i-t})}{\sum_{t}\exp{(g_{i-t})}},
  \end{aligned}
\end{equation}
where $\{\mathbf{V}^{g},\mathbf{W}^g,\mathbf{U}^g\}$ are learned parameters and $0\,{\leq}\,t\,{\leq}\,h$. As done in Equation~\ref{eq:inter_attn} except for the relative distance embedding, the decoder of \textsc{Gate} also attends over all the encoder hidden states. And the question-level importance $\bar{g}_{i-t}$ is employed as the coefficient of the attention scores at turn $i\!-\!t$.

\subsection{Precedent SQL as Context}\label{sec:sql_history}

\begin{figure}[t]
    \centering
    \includegraphics[width=.85\linewidth]{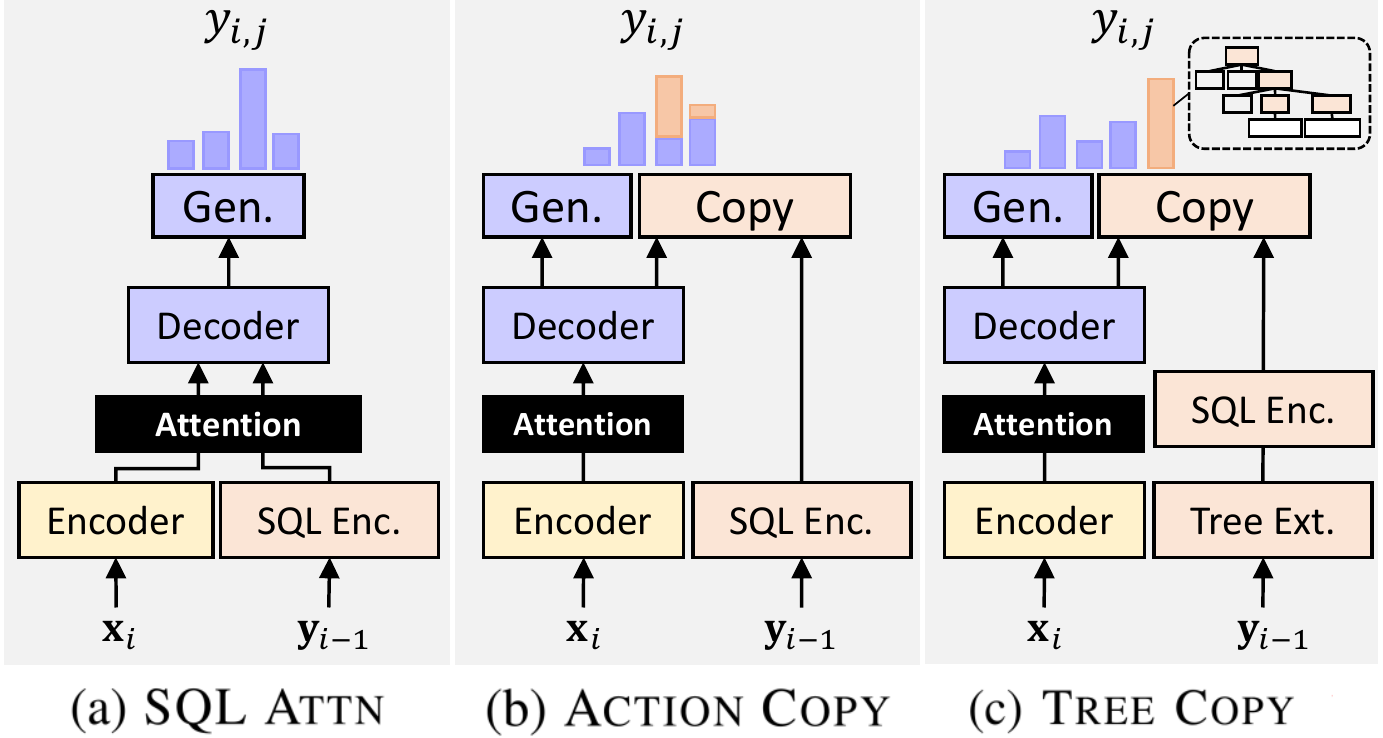}
    \caption{Different methods to employ the precedent SQL $\mathbf{y}_{i-1}$. \textit{SQL Enc.} is short for SQL Encoder, \textit{Tree Ext.} for Subtree Extractor, and \textit{Gen.} for Generate. (a) \textsc{SQL Attn}: attending over $\mathbf{y}_{i-1}$; (b) \textsc{Action Copy}: allow to copy actions from $\mathbf{y}_{i-1}$; (c) \textsc{Tree Copy}: allow to copy action subtrees extracted from $\mathbf{y}_{i-1}$.}
    \label{fig:diff_sql_context}
\end{figure}

Besides recent questions, as mentioned in Section~\ref{sec:intro}, the precedent SQL can also be context. As shown in Figure~\ref{fig:diff_sql_context}, the usage of $\mathbf{y}_{i-1}$ requires a SQL encoder, where we employ another BiLSTM to achieve it. The $m$-th contextual action representation at turn $i\!-\!1$, $\mathbf{h}^A_{i-1,m}$, can be obtained by passing the action sequence through the SQL encoder.

\paragraph{\textsc{SQL Attn}.} Attention over $\mathbf{y}_{i-1}$ is a straightforward method to incorporate the SQL context. Given $\mathbf{h}^A_{i-1,m}$, we employ a similar manner as Equation~\ref{eq:attn} to compute attention score and thus obtain the SQL context vector. This vector is employed as an additional input for decoder in Equation~\ref{eq:decoder}.

\paragraph{\textsc{Action Copy}.} To reuse the precedent generated SQL, \citet{zhang-etal-2019-editing} presented a token-level copy mechanism on their non-grammar based parser. Inspired by them, we propose an action-level copy mechanism suited for grammar-based decoding. It enables the decoder to copy actions appearing in $\mathbf{y}_{i-1}$, when the actions are compatible to the current expanded nonterminal. As the copied actions lie in the same semantic space with the generated ones, the output probability for action $\gamma$ is a mix of generating ($\mathbf{g}$) and copying ($\mathbf{c}$). The generating probability $P(y_{i,j}\!=\!{\gamma}\,|\,\mathbf{g})$ follows Equation~\ref{eq:global_output} and \ref{eq:schema_output}, while the copying probability is:
\begin{equation}
    \resizebox{.85\hsize}{!}{$\!\!P(y_{i,j}\!=\!{\gamma}|\mathbf{c})\,{\propto}\sum\nolimits_{m}\!\mathbf{1}[{\gamma\!=\!y_{i-1,m}}]\!\cdot\exp(\mathbf{h}^{\overrightarrow{D}}_{i,j}\mathbf{W}^l\mathbf{h}^A_{i-1,m})$},
\end{equation}
where $\mathbf{W}^l$ is a learned matrix. Denoting $P^{copy}_{i,j}$ the probability of copying at decoding step $j$ of turn $i$, it can be obtained by $\sigma(\mathbf{W}^{c}\mathbf{h}^{\overrightarrow{D}}_{i,j}+\mathbf{b}^{c})$, where $\{\mathbf{W}^{c},\mathbf{b}^{c}\}$ are learned parameters and $\sigma$ is the sigmoid function. The final probability $P(y_{i,j}={\gamma})$ is computed by:
\begin{equation}
    P^{copy}_{i,j}\,{\cdot}\,P(y_{i,j}={\gamma}\,|\,\mathbf{c})+(1-P^{copy}_{i,j})\,{\cdot}\,P(y_{i,j}={\gamma}\,|\,\mathbf{g}).
\end{equation}

\paragraph{\textsc{Tree Copy}.} Besides the action-level copy, we also introduce a tree-level copy mechanism. As illustrated in Figure~\ref{fig:diff_sql_context}, tree-level copy mechanism enables the decoder to copy action subtrees extracted from $\mathbf{y}_{i-1}$, which shrinks the number of decoding steps by a large margin. Similar idea has been proposed in a non-grammar based decoder \cite{suhr-etal-2018-learning}. In fact, a subtree is an action sequence starting from specific nonterminals, such as ${\rm Select}$. To give an example, $\langle$ \action{Select}{Agg}, \action{Agg}{max\ Col\ Tab}, \action{Col}{Id}, \action{Tab}{CARS\_DATA} $\rangle$ makes up a subtree for the tree in Figure~\ref{fig:model_overview}. For a subtree $\upsilon$, its representation $\phi^{t}(\upsilon)$ is the final hidden state when feeding its corresponding action sequence into the SQL encoder. Then we can obtain the output probability of subtree $\upsilon$ as:
\begin{equation}
    P(y_{i,j}={\upsilon})\ {\propto}\ \exp\big(\mathbf{h}^{\overrightarrow{D}}_{i,j}\mathbf{W}^t\phi^{t}(\upsilon)\big),
\end{equation}
where $\mathbf{W}^t$ is a learned matrix. The output probabilities of subtrees are normalized together with Equation~\ref{eq:global_output} and \ref{eq:schema_output}.

\subsection{BERT Enhanced Embedding}

We employ BERT \cite{devlin-etal-2019-bert} to augment our model via enhancing the embedding of questions and schemas. We first concatenate the input question and all the schemas in a deterministic order with \texttt{[SEP]} as delimiter \cite{hwang2019comprehensive}. For instance, the input for $Q_1$ in Figure~\ref{fig:context_intro} is ``What is id ... max horsepower? [SEP] CARS\_DATA ... [SEP] Make''. Feeding it into BERT, we obtain the schema-aware question representations and question-aware schema representations. These contextual representations are used to substitute $\phi^x$ subsequently, while other parts of the model remain the same.

\section{Experiment \& Analysis}\label{sec:exper}

We conduct experiments to study whether the introduced methods are able to effectively model context in the task of SPC (Section \ref{sec:quan}), and further perform a fine-grained analysis on various contextual phenomena (Section \ref{sec:ling}).

\subsection{Experimental Setup}

\paragraph{Dataset.} Two large complex cross-domain datasets are used: \textsc{SParC} \cite{yu-etal-2019-sparc} consists of $3034$\,/\,$422$ dialogues for train\,/\,development, and \textsc{CoSQL} \cite{yu-etal-2019-cosql} consists of $2164$\,/\,$292$ ones. The average turn numbers of \textsc{SParC} and \textsc{CoSQL} are $3.0$ and $5.2$, respectively. 

\paragraph{Evaluation Metrics.} We evaluate predicted SQL queries using exact set match accuracy \cite{yu-etal-2019-sparc}. Based on it, we consider three metrics: \textit{Question Match} (Ques.Match), the match accuracy over all questions, \textit{Interaction Match} (Int.Match), the match accuracy over all dialogues\footnote{Int.Match is much more challenging as it requires each predicted SQL in a dialogue to be correct.}, and \textit{Turn $i$ Match}, the match accuracy over questions at turn $i$. 

\paragraph{Implementation Detail.} Our implementation is based on PyTorch \cite{paszke2017automatic}, AllenNLP \cite{gardner2018allennlp} and the library transformers \cite{Wolf2019HuggingFacesTS}. We adopt the Adam optimizer and set the learning rate as $1$e-$3$ on all modules except for BERT, for which a learning rate of $1$e-$5$ is used \cite{kingma2014adam}. The dimensions of word embedding, action embedding and distance embedding are $100$, while the hidden state dimensions of question encoder, grammar-based decoder, turn-level encoder and SQL encoder are $200$. We initialize word embedding using GloVe \cite{pennington2014glove} for non-BERT models. For methods that use recent $h$ questions, $h$ is set as $5$ on both datasets.

\begin{table}[t]
    \centering
    \scalebox{0.78}{
        \begin{tabular}{lcccc}
        \toprule
           \multicolumn{1}{c}{\multirow{2}{*}{\textbf{Model}}} & \multicolumn{2}{c}{\textbf{SParC}} & \multicolumn{2}{c}{\textbf{CoSQL}}\\
         	\cmidrule(lr){2-3}
         	\cmidrule(lr){4-5}
             & {Ques.Match} & {Int.Match} & {Ques.Match} & {Int.Match}\\
             \midrule
             SyntaxSQL-con & $18.5$ & ~~$4.3$ & $15.1$ & ~~$2.7$\\
             CD-Seq2Seq & $21.9$ & ~~$8.1$ & $13.8$ & ~~$2.1$\\
             EditSQL & $33.0$ & $16.4$ & $22.2$ & ~~$5.8$ \\
             Ours & $\mathbf{41.8}$ & $\mathbf{20.6}$ & $\mathbf{33.5}$ & ~~$\mathbf{9.6}$\\
             \midrule
             EditSQL + BERT & $47.2$ & $29.5$ & $40.0$ & $11.0$ \\
             Ours + BERT & $\mathbf{52.6}$ & $\mathbf{29.9}$ & $\mathbf{41.0}$ & $\mathbf{14.0}$\\
        \bottomrule
        \end{tabular}
    }
    \caption{We report the best performance observed in 5 runs on the development sets of both \textsc{SParC} and \textsc{CoSQL}, since their test sets are not public. We also conduct Wilcoxon signed-rank tests between our method and the baselines, and the bold results show the improvements of our model are significant with p $<$ $0.005$.}
    \label{tab:experiment_result}
\end{table}

\paragraph{Baselines.} We consider three models as our baselines. SyntaxSQL-con and CD-Seq2Seq are two strong baselines introduced in the \textsc{SParC} dataset paper \cite{yu-etal-2019-sparc}. SyntaxSQL-con employs a BiLSTM model to encode dialogue history upon the SyntaxSQLNet model (analogous to our \textsc{Turn}) \cite{yu-etal-2018-syntaxsqlnet}, while CD-Seq2Seq is adapted from~\citet{suhr-etal-2018-learning} for cross-domain settings (analogous to our \textsc{Turn+Tree Copy}). EditSQL \cite{zhang-etal-2019-editing} is a STOA baseline which mainly makes use of SQL attention and token-level copy (analogous to our \textsc{Turn+SQL Attn+Action Copy}).

\subsection{Model Comparison}\label{sec:quan}

\begin{figure}
    \centering
    \includegraphics[width=\linewidth]{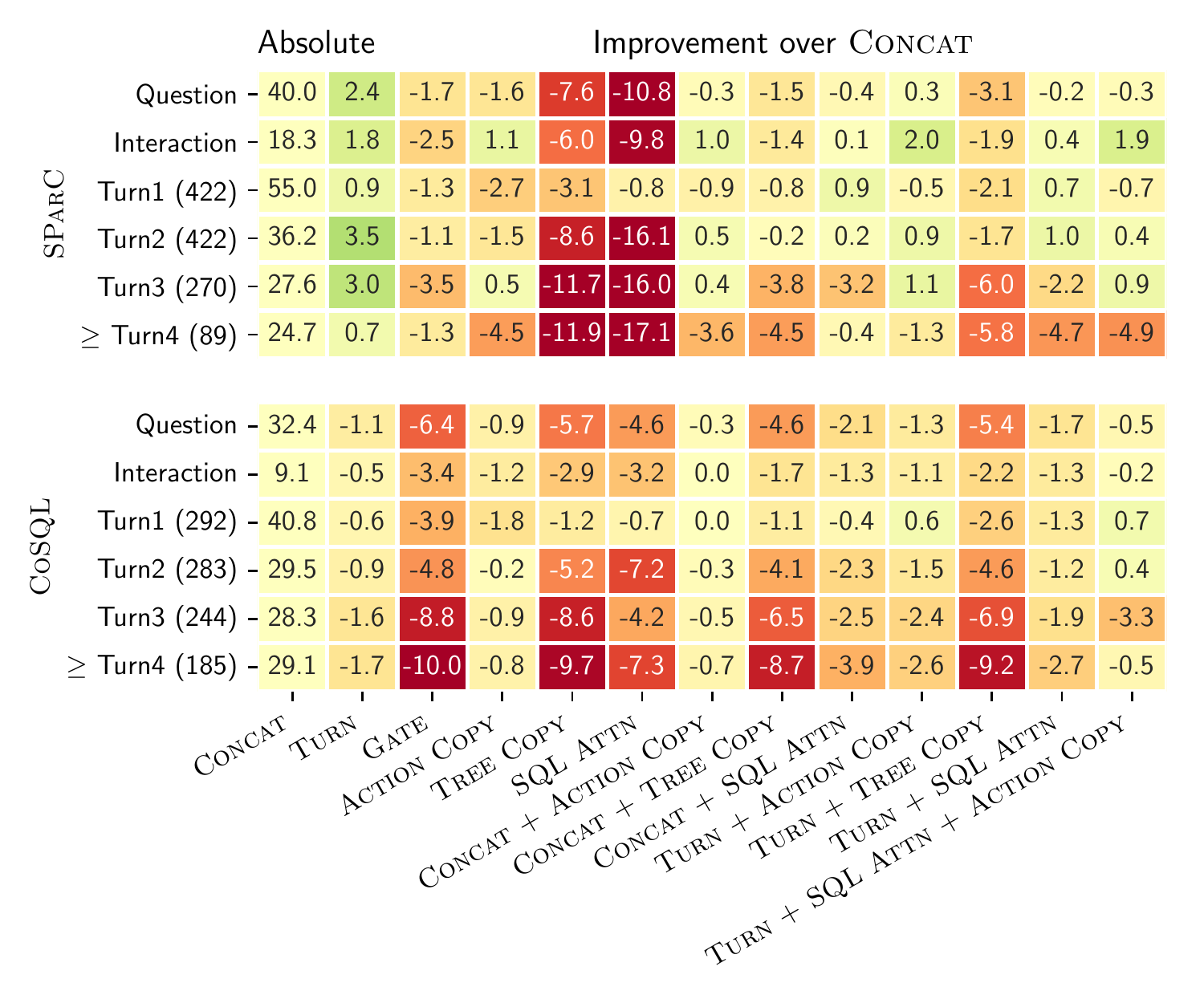}
    \caption{Question Match, Interaction Match and Turn $i$ Match on \textsc{SParC} and \textsc{CoSQL} development sets. The numbers are averaged over 5 runs. The first column represents \textit{absolute} values. The rest are \textit{improvements} of different context modeling methods over \textsc{Concat}.}
\label{fig:performance_dev_overall}
\end{figure}

Taking \textsc{Concat} as a representative, we compare the performance of our model with other models, as shown in Table~\ref{tab:experiment_result}. As illustrated, our model outperforms baselines by a large margin with or without BERT, achieving new SOTA performances on both datasets. Compared with the previous SOTA without BERT on \textsc{SParC}, our model improves Ques.Match and Int.Match by $8.8$ and $4.2$ points, respectively.

To conduct a thorough comparison, we evaluate $13$ different context modeling methods upon the same parser, including $6$ methods introduced in Section~\ref{sec:method} and $7$ selective combinations of them (e.g., \textsc{Concat+Action Copy}). The experimental results are presented in Figure~\ref{fig:performance_dev_overall}. Taken as a whole, it is very surprising to observe that none of these methods can be consistently superior to the others. The experimental results on BERT-based models show the same trend. Diving deep into the methods only using recent questions as context, we observe that \textsc{Concat} and \textsc{Turn} perform competitively, outperforming \textsc{Gate} by a large margin. With respect to the methods only using precedent SQL as context, \textsc{Action Copy} significantly surpasses \textsc{Tree Copy} and \textsc{SQL Attn} in all metrics. In addition, we observe that there is little difference in the performance of \textsc{Action Copy} and \textsc{Concat}, which implies that using precedent SQL as context gives almost the same effect with using recent questions. In terms of the combinations of different context modeling methods, they do not significantly improve the performance as we expected. 

\begin{figure}[t]
    \centering
    \includegraphics[width=.75\linewidth]{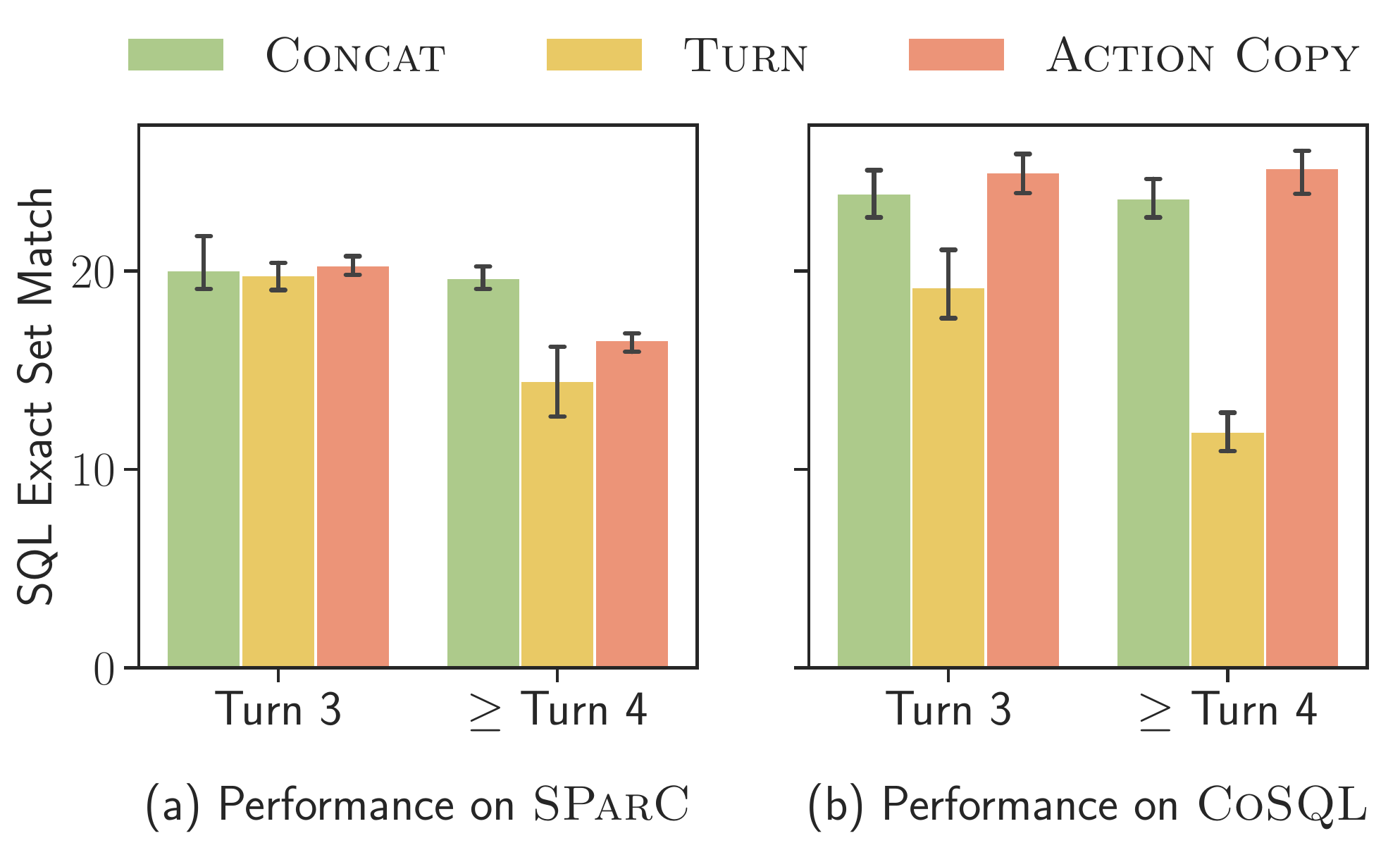}
    \caption{Out-of-distribution experimental results (Turn $i$ Match) of three models on \textsc{SParC} and \textsc{CoSQL} development sets.}
    \label{fig:out_dis_result}
\end{figure}

As mentioned in Section~\ref{sec:intro}, intuitively, methods which only use the precedent SQL enjoy better generalizability. To validate it, we further conduct an \textit{out-of-distribution} experiment to assess the generalizability of different context modeling methods. Concretely, we select three representative methods and train them on questions at turn $1$ and $2$, whereas test them at turn $3$, $4$ and beyond. As shown in Figure~\ref{fig:out_dis_result}, \textsc{Action Copy} has a consistently comparable or better performance, validating the intuition. Meanwhile, \textsc{Concat} appears to be strikingly competitive, demonstrating it also has a good generalizability. Compared with them, \textsc{Turn} is more vulnerable to out-of-distribution questions. In conclusion, existing context modeling methods in the task of SPC are not as effective as expected, since they do not show a significant advantage over the simple concatenation method.

\begin{table*}[t]
    \centering
    \scalebox{0.75}{
        \begin{tabular}{cllrll}
        \toprule
             \multicolumn{1}{l}{\multirow{2}{*}{\textbf{Contextual Phenomena}}} & \multicolumn{2}{l}{\multirow{2}{*}{\textbf{Fine-grained Types}}} &
             \multicolumn{1}{l}{\multirow{2}{*}{\textbf{Count}}} &
             \multicolumn{2}{c}{\textbf{Example}}\\
         & & & & \multicolumn{1}{c}{Precedent Question} & \multicolumn{1}{c}{Current Question} \\
             \midrule
             Semantically Complete & \multicolumn{2}{l}{Context Independent} & $149$ & Show the nationality of each person. & Group people by their nationality.\\
             \midrule
            \multirow{5}{*}{Coreference} & \multicolumn{2}{l}{ Bridging Anaphora} & $31$ & Show the version number for all templates. & What is the smallest \underline{value}? \\
             & \multicolumn{2}{l}{Definite Noun Phrases} & $67$ & Which country has a head of state named Beatrix?  & What languages are spoken in \underline{that country}?\\
             & \multicolumn{2}{l}{One Anaphora} & $59$ & Order the pets by age. & How much does each \underline{one} weigh? \\
             & \multicolumn{2}{l}{Demonstrative Pronoun} & $195$ & Which students have pets? & Of \underline{those}, whose last name is smith? \\
             & \multicolumn{2}{l}{Possessive Determiner} & $88$ & How many highschoolers are liked by someone else? & What are \underline{their} names? \\
             \midrule
              \multirow{5}{*}{Ellipsis} & \multicolumn{2}{l}{Continuation} & $131$ & What are all the flight numbers? & Which land in Aberdeen? \\
              & \multirow{4}{*}{Substitution} & Explicit & $61$ &  What is id of the car with the max horsepower? & How about with the max \doubleunderline{MPG}?\\
             & & Implicit & $60$ & Find the names of museums opened before 2010. & How about \doubleunderline{after}? \\
             & & Schema & $80$ & How many losers participated in the Australian Open? & \doubleunderline{Winners}?\\
              & & Operator & $41$ & Who was the last student to register? & Who was the \doubleunderline{first} to register? \\
        \bottomrule
        \end{tabular}
    }
    \caption{Different fine-grained types, their count and representative examples from the \textsc{SParC} development set. \underline{one} means \textit{one} is a pronoun. \doubleunderline{Winners} means \textit{Winners} is a phrase intended to substitute \textit{losers}.}
    \label{tab:phon}
\end{table*}

\subsection{Fine-grained Analysis}\label{sec:ling}

\begin{figure}[t]
    \centering
    \includegraphics[width=1\linewidth]{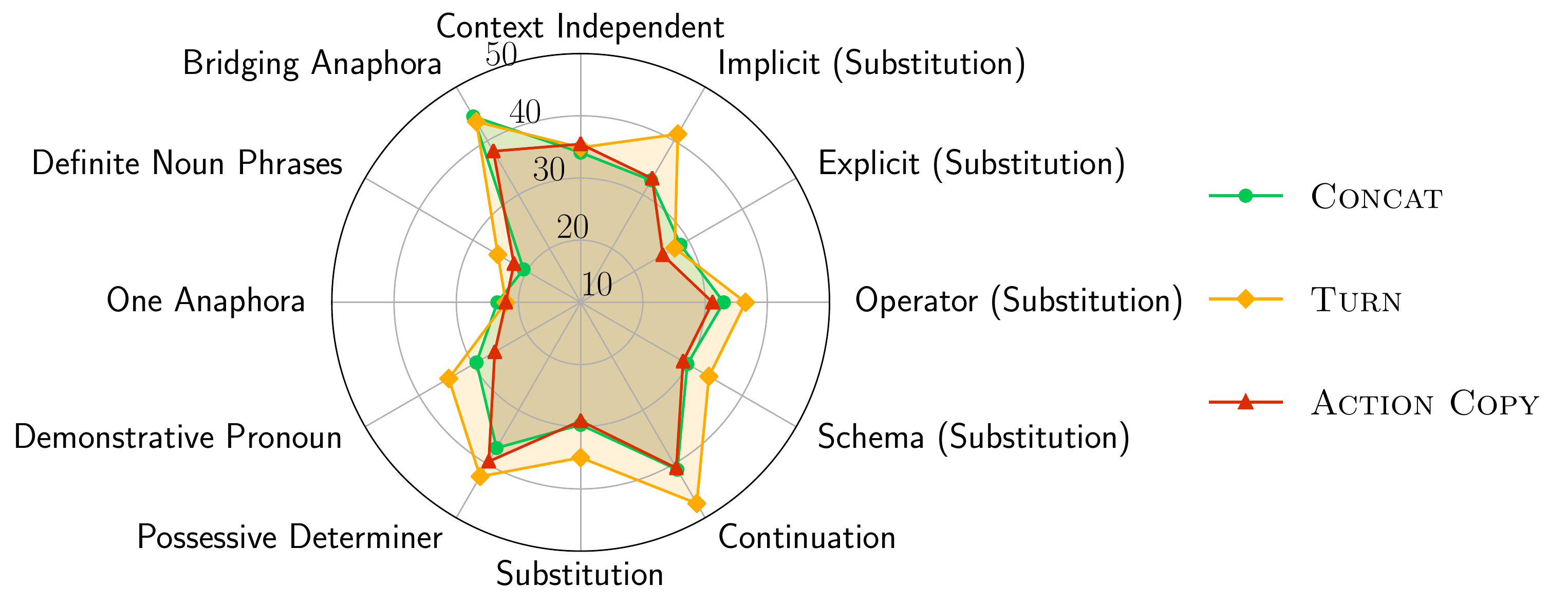}
    \caption{The average Ques.Match (across 5 runs) of different context modeling methods on fine-grained types. }
    \label{fig:radar_chart}
\end{figure}

By a careful investigation on contextual phenomena, we summarize them in multiple hierarchies. Roughly, there are three kinds of contextual phenomena in questions: \textit{semantically complete}, \textit{coreference} and \textit{ellipsis}. Semantically complete means a question can reflect all the meaning of its corresponding SQL. Coreference means a question contains pronouns, while ellipsis means the question cannot reflect all of its SQL, even if resolving its pronouns. In the fine-grained level, coreference can be divided into $5$ types according to its pronoun \cite{androutsopoulos1995natural}. Ellipsis can be characterized by its intention: \textit{continuation} and \textit{substitution}\footnote{The fine-grained types of ellipsis are proposed by us because there is no consensus yet.}. Continuation is to augment extra semantics (e.g. ${\rm Filter}$), and substitution refers to the situation where current question is intended to substitute particular semantics in the precedent question. Substitution can be further branched into $4$ types: \textit{explicit}\,vs.\,\textit{implicit} and \textit{schema}\,vs.\,\textit{operator}. Explicit means the current question provides contextual clues (i.e. partial context overlaps with the precedent question) to help locate the substitution target, while implicit does not. In most cases, the target is schema or operator. In order to study the effect of context modeling methods on various phenomena, as shown in Table~\ref{tab:phon}, we take the development set of \textsc{SParC} as an example to perform our analysis. The analysis begins by presenting Ques.Match of three representative models on the above fine-grained types in Figure~\ref{fig:radar_chart}. As shown, though different methods have different strengths, they all perform poorly on certain types, which will be elaborated below.

\paragraph{Coreference.} Diving deep into the coreference (left of Figure~\ref{fig:radar_chart}), we observe that all methods struggle with two fine-grained types: \textit{definite noun phrases} and \textit{one anaphora}. Through our study, we find the scope of antecedent is a key factor. An antecedent is one or more entities referred by a pronoun. Its scope is either \textit{whole}, where the antecedent is the precedent answer, or \textit{partial}, where the antecedent is part of the precedent question. The above-mentioned fine-grained types are more challenging as their partial proportion are nearly $40\%$, while for \textit{demonstrative pronoun} it is only $22\%$. It is reasonable as partial requires complex inference on context. Considering the $4^{\text{th}}$ example in Table~\ref{tab:phon}, ``one'' refers to ``pets'' instead of ``age'' because the accompanying verb is ``weigh''. From this observation, we draw the conclusion that current context modeling methods do not succeed on pronouns which require complex inference on context.

\paragraph{Ellipsis.} As for ellipsis (right of Figure~\ref{fig:radar_chart}), we obtain some findings by comparisons in three aspects. The first finding is that all models have a better performance on continuation than substitution. This is expected since there are redundant semantics in substitution, while not in continuation. Considering the $8^{\text{th}}$ example in Table~\ref{tab:phon}, ``horsepower'' is redundant and it may raise noise in SQL prediction. The second finding comes from the unexpected drop from \textit{implicit(substitution)} to \textit{explicit(substitution)}. Intuitively, explicit should surpass implicit on substitution as it provides more contextual clues. The finding demonstrates that contextual clues are obviously not well utilized by the context modeling methods. Third, compared with \textit{schema(substitution)}, \textit{operator(substitution)} achieves a comparable or better performance consistently. We believe it is caused by the cross-domain setting, which makes schema related substitution more difficult.

\section{Related Work}

The most related work is the line of semantic parsing in context. In the topic of SQL, \citet{zettlemoyer2009learning} proposed a context-independent CCG parser and then applied it to do context-dependent substitution, \citet{iyyer-etal-2017-search} applied a search-based method for sequential questions, and \citet{suhr-etal-2018-learning} provided the first sequence-to-sequence solution in the area. More recently, \citet{zhang-etal-2019-editing} presented a edit-based method to reuse the precedent generated SQL, while \citet{liu-etal-2019-split} introduced an auxiliary task. With respect to other logic forms, \citet{long-etal-2016-simpler} focused on understanding execution commands in context, \citet{guo2018dialog} on question answering over knowledge base in a conversation, and \cite{iyer-etal-2018-mapping} on code generation in environment context. Our work is different from theirs as we perform an exploratory study, not fulfilled by previous works.

There are also several related works that provided studies on context. \citet{hwang2019comprehensive} explored the contextual representations in context-independent semantic parsing, and \citet{sankar-etal-2019-neural} studied how conversational agents use conversation history to generate response. Different from them, our task focuses on context modeling for semantic parsing. Under the same task, \citet{androutsopoulos1995natural} summarized contextual phenomena in a coarse-grained level, while \citet{bertomeu-etal-2006-contextual} performed a Wizard-of-Oz experiment to study the most frequent phenomena. What makes our work different from them is that we not only summarize contextual phenomena by fine-grained types, but also perform an analysis of context modeling methods.

\section{Conclusion \& Future Work}

This work conducts an exploratory study on semantic parsing in context, to realize how far we are from effective context modeling. Through a thorough comparison, we find that existing context modeling methods are not as effective as expected. A simple concatenation method can be much competitive. Furthermore, by performing a fine-grained analysis, we summarize two potential directions as our future work: incorporating common sense for better pronouns inference, and modeling contextual clues in a more explicit manner. We believe our work can facilitate the community to debug models in a fine-grained level and make more progress.



\bibliographystyle{named}
\bibliography{ijcai20}

\end{document}